\documentclass{article}
\usepackage{spconf,amsmath,graphicx,multirow}
\usepackage{hyperref}
\usepackage{amssymb,xcolor}

\usepackage{enumitem}
\setlist{nosep, leftmargin=14pt}

\usepackage{mwe} 

\title{QwenCLIP: Boosting Medical Vision-Language Pretraining via LLM Embeddings and Prompt tuning}
\name{Xiaoyang Wei \qquad Camille Kurtz \qquad Florence Cloppet}
\address{
    Laboratoire d'Informatique Paris Descartes (LIPADE), Université Paris Cité (France)
}
\begin{document}
%
\maketitle
\begin{abstract}
Contrastive Language–Image Pretraining (CLIP) has demonstrated strong generalization for vision-language tasks in computer vision and medical domains, yet its text encoder accepts only up to 77 tokens, which limits its ability to represent long and information-rich radiology reports. Recent adaptations using domain-specific encoders, such as PubMedBERT or ClinicalBERT, mitigate this issue by leveraging medical corpora, but remain constrained by their limited input length (typically 512 tokens) and relatively shallow semantic understanding. 
To address these limitations, we propose QwenCLIP, a vision-language framework that replaces CLIP’s text encoder with a large language model (LLM)–based embedding module (e.g., Qwen3-Embedding) and introduces learnable prompts to enhance cross-modal alignment. 
By leveraging the extended context window and richer representations of LLMs, QwenCLIP captures comprehensive medical semantics from long-form clinical text, substantially improving medical image–text alignment and downstream performance on radiology benchmarks.
Our code is publicly available at \href{https://github.com/Wxy-24/QwenCLIP}{https://github.com/Wxy-24/QwenCLIP}.

\end{abstract}
\section{Introduction}
\label{sec:intro}


The advent of Contrastive Language–Image Pretraining (CLIP) \cite{radford2021learning} has revolutionized medical visual representation learning, enabling models to acquire rich multimodal understanding through self-supervised learning rather than relying on exhaustive manual annotation. Instead of using expensive and time-consuming expert annotations for medical images, researchers are increasingly leveraging abundant image–text pairs readily available in clinical and public data sources. 
These pairs typically originate from two major domains: (1) radiological images with corresponding diagnostic reports and clinical notes routinely stored in Picture Archiving and Communication Systems (PACS) \cite{irvin2019chexpert}, and (2) medical figures and captions extracted from publicly available biomedical literature \cite{ROCOV2}. 
Such resources offer scalable and diverse supervision signals for aligning visual and textual representations in the medical domain \cite{wei2024integrating}.

\begin{figure}[!t]
\begin{center}
\includegraphics[width=1.0\linewidth]{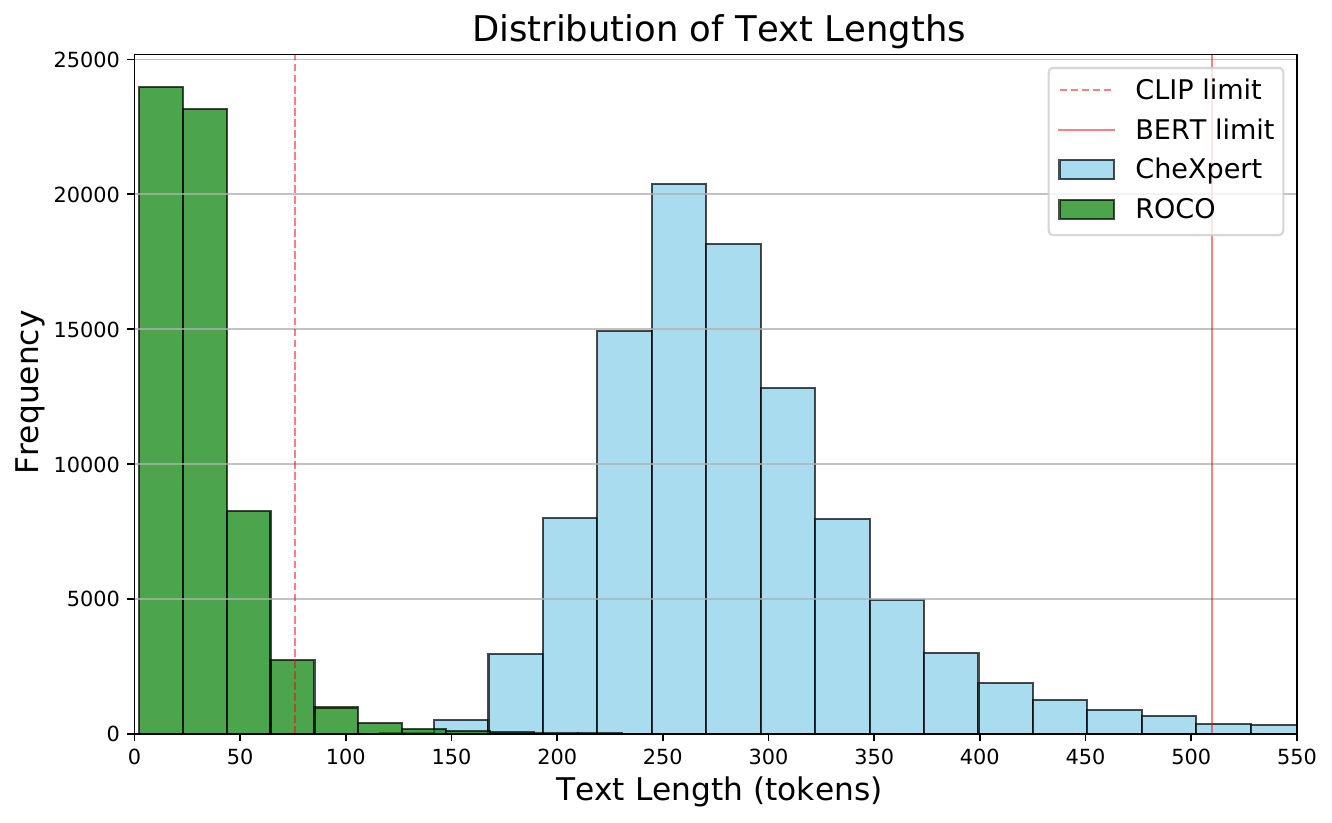}
\end{center}
\caption{Text length distribution for common medical image-text pair datasets.
}
\label{fig:distribution}
\vspace{-4mm}
\end{figure}

Although CLIP has achieved remarkable success by utilizing large-scale, freely available image–text pairs, it still presents several non-negligible limitations when directly applied to medical data. One major constraint lies in its text encoder, which accepts no more than 77 tokens—a design choice suitable for short, web-scraped captions (e.g., “a cat sitting on a sofa”), but clearly insufficient for long and information-dense medical texts. Radiology reports, for instance, often exceed several hundred tokens, containing detailed descriptions of anatomical structures, imaging findings, and diagnostic impressions. When fed into CLIP, such text must be truncated, resulting in inevitable information loss and degraded alignment with the corresponding medical image.

To better capture domain-specific semantics, several studies have proposed replacing CLIP’s text encoder with medical language models such as PubMedBERT \cite{pubmedbert} or ClinicalBERT \cite{clinicalbert}, which are pretrained on large-scale biomedical corpora. These models indeed improve text understanding by incorporating medical terminology and contextual nuances. However, BERT-based encoders inherently support a limited context window (up to 512 tokens), which remains inadequate for processing long-form clinical narratives—especially multi-section radiology reports or comprehensive case descriptions (as illustrated in Fig.~\ref{fig:distribution}). Thus, while domain-specific BERT variants alleviate part of the problem, they cannot fully address the context length bottleneck.

In contrast, large language models (LLMs) have recently demonstrated extraordinary capabilities in handling long textual contexts and capturing rich semantic relationships across sentences and paragraphs \cite{li2025generation}. 
Furthermore, LLMs are typically pretrained on vast, diverse corpora spanning the entire internet, allowing them to benefit from scaling laws and acquire a broad base of intrinsic knowledge. This often enables them to outperform BERT-like encoders that are trained on more limited domain-specific data. 
For example, Gemini 2.5 Pro can process up to 1 million tokens, such technological advancement can effectively eliminate the concern of the input-length limitations faced by BERT and the original CLIP encoder. This ability is particularly advantageous for modeling detailed medical reports, where long radiology descriptions frequently exceed conventional context windows and risk truncation.
Consequently, employing an LLM-based text encoder provides a promising avenue for enhancing medical vision–language alignment.

Recent work such as LLM2CLIP \cite{huang2024llm2clip} has explored integrating general-purpose LLMs (e.g., LLaMA-based models \cite{dubey2024llama}) into the CLIP framework, demonstrating the potential of large language models but introduce significant complexities: they require extensive two-stage training pipelines where generative LLMs must first be converted into discriminative encoders through caption contrastive fine-tuning. This not only increases computational overhead but also represents a fundamental architectural mismatch—adapting text generation models for embedding tasks.
In contrast, we propose QwenCLIP, which fundamentally rethinks this integration strategy. Rather than repurposing generative LLMs through complex adaptation pipelines, it directly employs Qwen3-embedding \cite{zhang2025qwen3}—a dedicated embedding model specifically designed for representation learning. This eliminates the need for the cumbersome pre-processing stages required by methods like LLM2CLIP while providing native support for long contexts.
Furthermore, recognizing the sensitivity of LLMs to input phrasing and prompts, QwenCLIP introduces a hybrid prompt tuning strategy that enables the model to autonomously learn task-adaptive prompts for effective cross-modal alignment. Together, these innovations allow QwenCLIP to bridge the gap between long-text medical understanding and visual representation learning, providing an efficient, scalable, and semantically rich solution for medical vision–language modeling.

\begin{figure}[!t]

\begin{center}
\includegraphics[width=1.0\linewidth]{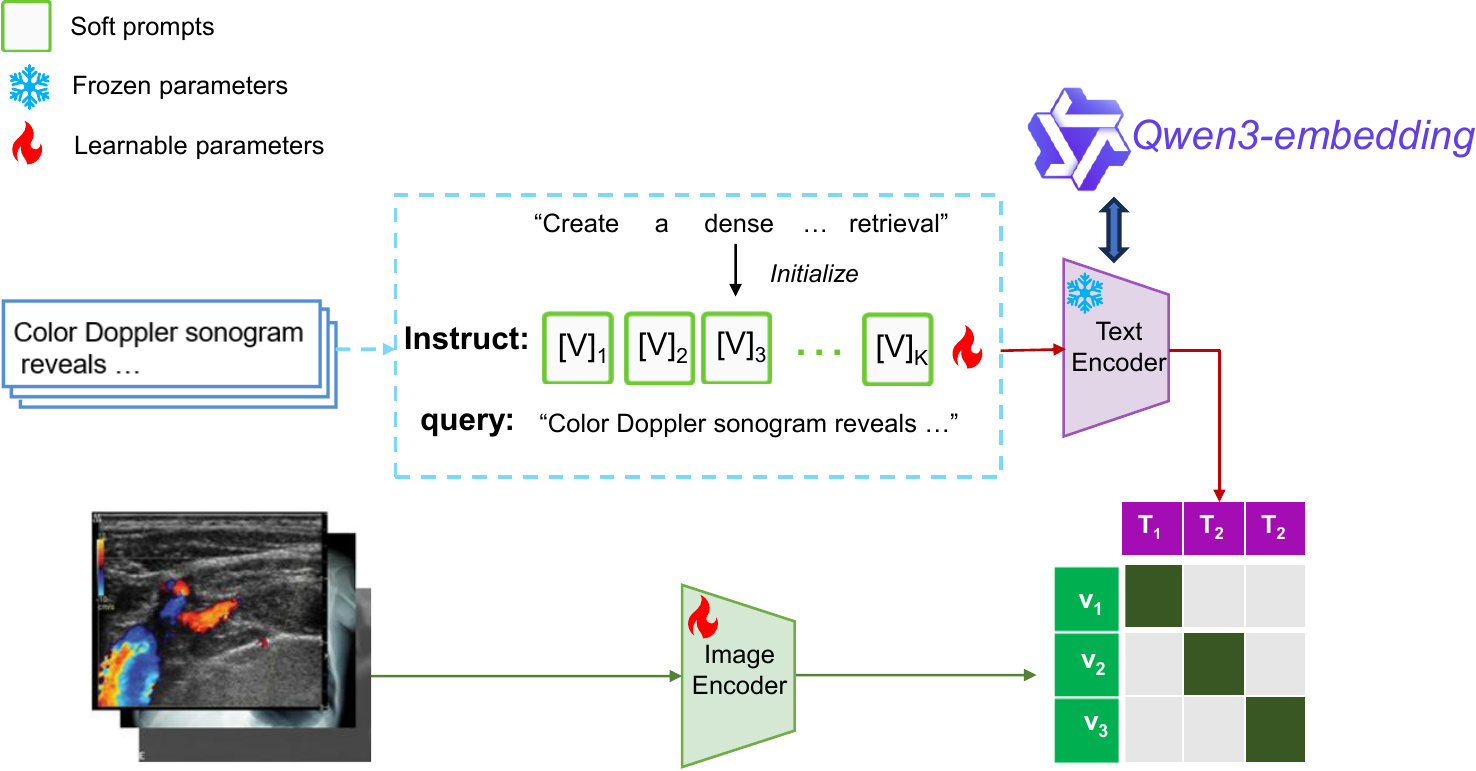}
\end{center}
\vspace{-3mm}
\caption{Overview of the QwenCLIP framework, where instruction-tuned soft prompts guide the frozen LLM and learnable image encoder to align multimodal representations}
\label{fig:workflow}
\vspace{-4mm}
\end{figure}

\section{Proposed framework: QwenCLIP}

Given a dataset of paired medical images and captions, our \textbf{QwenCLIP} framework aims to learn semantically aligned visual embeddings for downstreams tasks like image retrieval. 
The method extends the original CLIP architecture by (1) replacing its text encoder with the \textbf{Qwen3-embedding-8B} model \cite{zhang2025qwen3},
which is a high-capacity LLM optimized for dense text representations, and (2) introducing a \textbf{hybrid prompt tuning strategy} to better capture medical contextual information (as illustrated in Fig.~\ref{fig:workflow}).



\subsection{Hybrid prompt tuning for LLM-based text encoder}

Traditional CLIP often employs a fixed template (e.g., ``a photo of a $\ldots$'') for zero-shot classification. 
However, such handcrafted prompts tend to be inadequate for medical contexts where textual descriptions are domain-specific and context-dependent. 
Inspired by CoOp \cite{zhou2022learning}, we design a \textbf{hybrid prompting strategy} that combines \textit{static explicit prompts} with a \textit{learnable contextual soft prompt}.

Specifically, we prepend the input caption with explicit prefixes such as 
\textbf{``Instruct:''} and \textbf{``query:''} to provide structural cues to the LLM about the intended task (instructional vs. retrieval-oriented semantics). 
In addition to these static tokens, we introduce a \textbf{learnable soft prompt vector} $P_s$ designed with the following form:,
\[
P_s = [V]_1[V]_2\dots[V]_K.
\]
 where each $[V]_k \ (k \in \{1, \dots, K\})$  is a learnable vector with the same dimension as word embeddings, and $K$ is a hyperparameter corresponding to the number of context tokens which depend on the initialization text. Specifically, we initialized $P_s$ using the following phrase with $K=15$ in this work:
\begin{quote}
\textit{``Create a dense embedding that represents the medical meaning of this text for image retrieval.''}
\end{quote}
During training, $P_s$ is optimized jointly with the model parameters, allowing the text encoder to dynamically refine the contextualized embedding space while preserving the linguistic priors of the pretrained Qwen3-embedding model. Formally, given a caption $T$, we construct the final textual input as:
\[
T' = [\text{Instruct: $P_s$,  query:}  T].
\]
where \textbf{``Instruct:''} and \textbf{``query:''} refer to the static explicit prompt during training. Finally, the Qwen3-embedding model encodes $T'$ to yield the text feature $t = H_{\text{text}}(T')$.

\subsection{Vision-language contrastive learning}

The image encoder $H_{\text{visual}}$ is a Vision Transformer (ViT-B/16) initialized with CLIP weights. Given an image-text pair $(I, T)$, we obtain their embeddings:
\[
v = H_{\text{visual}}(I), \quad t = H_{\text{text}}(T').
\]
To ensure both embeddings are projected into a shared latent space, we keep the original design of the image encoder and append a two-layer MLP to the LLM encoder to adapt its high dimensional dense embedding.
Following CLIP, we train the model using the \textbf{InfoNCE} loss in a symmetric contrastive manner, which encourages matched image--text pairs to have higher similarity than mismatched ones.

The image-to-text and text-to-image contrastive losses are respectively defined as:
\[
\mathcal{L}_{i2t} = -\frac{1}{N}\sum_{i=1}^{N} \log \frac{\exp(\text{sim}(v_i, t_i)/\tau)}{\sum_{j=1}^{N} \exp(\text{sim}(v_i, t_j)/\tau)},
\]
\[
\mathcal{L}_{t2i} = -\frac{1}{N}\sum_{i=1}^{N} \log \frac{\exp(\text{sim}(t_i, v_i)/\tau)}{\sum_{j=1}^{N} \exp(\text{sim}(t_i, v_j)/\tau)},
\]
where $\text{sim}(\cdot,\cdot)$ denotes the cosine similarity and $\tau$ is a learnable temperature parameter. The final training objective is the symmetric InfoNCE loss:
\[
\mathcal{L}_{\text{CLIP}} = \mathcal{L}_{i2t} + \mathcal{L}_{t2i}.
\]

This objective aligns medical image representations with semantically enriched text embeddings produced by the hybrid-prompted Qwen3 encoder, enabling robust cross-modal understanding even under domain-specific variations in terminology and context.

\begin{table*}[!t]
\caption{Zero-shot image-to-image retrieval performance on ROCOv2 (CUI@K, P@K) and IRMA (P@K) using ViT-B/16 as the image encoder with various baselines and their text encoders shown in the left columns.
}
\resizebox{\textwidth}{18mm}{
{\small
\setlength{\tabcolsep}{3pt}
\begin{tabular}{|l|c|rrr|rrr|rrr|rrr|rrr|}
\hline
\multirow{3}{*}{Methods} &\multirow{3}{*}{Text encoder (-params)}& \multicolumn{3}{c|}{ROCOv2} & \multicolumn{9}{c|}{Custom retrieval dataset (ROCOv2)} & \multicolumn{3}{c|}{IRMA} \\
 && \multicolumn{3}{c|}{CUI@K} & \multicolumn{9}{c|}{P@K} & \multicolumn{3}{c|}{P@K} \\
 && @5 & @10 & @50 & @5 & @10 & @50 & @5 & @10 & @50 & @5 & @10 & @50 & @5 & @10 & @50 \\ \hline
 && & & & \multicolumn{3}{c|}{Modality} & \multicolumn{3}{c|}{Organ} & \multicolumn{3}{c|}{Modality \& Organ} & \multicolumn{3}{c|}{Organ} \\ 
CLIP \cite{radford2021learning} &GPT2-117M& 44.64   & 45.90  & 50.53 & 94.41 & 92.66 & 90.58& 83.70& 80.38 & 74.31 & 89.11     & 87.66    & 83.96  & 97.08 & 97.05 & 96.18 \\
ClinicalCLIP \cite{serieys2022text}&ClinicalBERT-110M& 45.18    & 46.62  & 50.97  & 94.63 & 92.91 & 90.91 & 83.83 & 80.44 & 74.11 &   89.25    &  87.78    &  83.93 &   98.03    &  97.26    &  96.45 \\
PMC-CLIP \cite{lin2023pmc}&PubMedBERT-340M& 45.58    & 46.93  & 51.71  & 94.98 & 92.88 & 90.96 & 85.69 & 83.87 & 78.01 &  90.72 & 88.91 &  86.93 &   98.13    &  97.32    &  96.48 \\
LLM2CLIP \cite{huang2024llm2clip}&Llama-3-Instruct-8B& 45.43 & 46.72 & 51.78 & 96.00 & 94.41 & 92.27 & 85.40 & 83.14 & 78.63 & 90.88 & 90.31 & 88.02 & 98.15 & 97.29 & 96.59 \\
\hline
QwenCLIP &Qwen3-embedding-8B& \textbf{45.96} & \textbf{47.23} & \textbf{52.47} & \textbf{96.24} & \textbf{94.78} & \textbf{92.61} & \textbf{85.99} & \textbf{83.72} & \textbf{79.22} & \textbf{91.35} & \textbf{90.69} & \textbf{88.42} & \textbf{98.49} & \textbf{97.67} & \textbf{97.34} \\
\hline


\end{tabular}}}
\label{tab1}
\end{table*}

\section{Experimental study}
\label{sec:exp}

We present in this section an experimental study to compare the effectiveness of our QwenCLIP framework and state-of-the-art alternatives which aim to improve vision-language pretraining using various text encoders. 
Since we freeze our text encoder during pretraining, we propose to evaluate the learned visual representation under image-to-image retrieval (downstream) tasks.

\subsection{Datasets}
\textbf{ROCOv2} \cite{ROCOV2} is an extended multimodal medical image dataset introduced with all images extracted from PubMed Central open-access articles along with their corresponding figure captions, encompassing a broad spectrum of imaging modalities and anatomical regions. 
It consists of 60163, 9945, and 9972 images in the training, validation, and testing splits, respectively. Besides, it is created using improved caption cleaning and de-duplication procedures to enhance text quality. While the average caption length in ROCOv2 is 32 tokens, 4.4\% of them exceed CLIP's 77-token limit. We employ ROCOv2 both for pre-training and evaluation.

\textbf{IRMA} \cite{lehmann2003irma} is an X-ray image dataset of 14k images of different parts of the human body. Each image is associated with a 13-digit hierarchical IRMA code specifying annotations in terms of modality, direction and anatomy. 
We evaluated image retrieval performance based on the IRMA code. 



\subsection{Implementation details}
\label{ssec:task}
We adopted the official training validation split of ROCOv2. 
All images were resized to size of 224 $\times{}$ 224. 
We initialized ViT- B/16 as image encoder using \href{https://openaipublic.azureedge.net/clip/models/5806e77cd80f8b59890b7e101eabd078d9fb84e6937f9e85e4ecb61988df416f/ViT-B-16.pt}{default CLIP weights}. 
\href{https://huggingface.co/Qwen/Qwen3-Embedding-8B}{Qwen3-embedding-8B} is loaded as the text encoder. 
To reduce memory usage, we freeze the whole text encoder and computed the feed-forward pass of text embedding using half precision (float16). 
The model was trained on a NVIDIA A100 80GB GPU with a batch size of 32 for 10 epochs. 
We used the Adam optimizer with a learning rate set to 3e-6. 

\subsection{Downstream tasks}
 We evaluate our methods with zero shot image-to-image retrieval tasks using the evaluation metrics below.
 
\textbf{CUI@K} is an evaluation metric based on Concept Unique Identifiers (CUIs) proposed in \cite{serieys2022text}. 
Specifically, for each query image, we calculate the Intersection over Union (IoU) between its associated set of CUIs and those of all candidate images. The candidates are then ranked in descending order of their IoU values, and this ranking serves as the ground truth for the image-to-image retrieval task.
Subsequently, we compute the CUI@K metric using the Normalized Discounted Cumulative Gain (NDCG) formulation, where $K$ denotes the number of top retrieved images. The final performance measure is obtained by averaging the CUI@K values across all query images.

\textbf{Precision@K} is a standard metric in information retrieval, computed as the proportion of relevant samples among the top-$K$ retrieved results. 
Here, image relevance is defined based on shared categorical attributes such as modality, organ, or imaging direction. To this end, we construct a customized retrieval benchmark from ROCOv2 (referred to as the “custom retrieval dataset”). Specifically, we derive modality and organ labels using the semantic types of CUIs: images associated with the semantic type T060 (“diagnostic procedure”) are grouped into five imaging modalities (e.g., ultrasound, plain X-ray, tomography), while those with T023 (“body part, organ”) are assigned to ten organ categories (e.g., stomach, teeth, lung). Using these labels, we define three retrieval tasks—\textit{Modality}, \textit{Organ}, and \textit{Modality \& Organ}—and evaluate model performance using Precision@K.

\begin{table*}[!t]
\caption{Ablation study of QwenCLIP framework for two main contributions in this work. Here "LLM" refers to replacing original text encoder of CLIP with Qwen3-embedding 8B, "LP" stands for applying learnable prompt for textual input of LLMs.}
\setlength{\tabcolsep}{3.5pt}
{\small
\begin{tabular}{|l|cc|rrr|rrr|rrr|rrr|rrr|}
\hline
\multirow{3}{*}{Methods} &\multirow{3}{*}{LLM}&\multirow{3}{*}{LP}& \multicolumn{3}{c|}{ROCOv2} & \multicolumn{9}{c|}{Custom retrieval dataset} & \multicolumn{3}{c|}{IRMA} \\
 &&& \multicolumn{3}{c|}{CUI@K} & \multicolumn{9}{c|}{P@K} & \multicolumn{3}{c|}{P@K} \\
 &&& @5 & @10 & @50 & @5 & @10 & @50 & @5 & @10 & @50 & @5 & @10 & @50 & @5 & @10 & @50 \\ \hline
 &&& & & & \multicolumn{3}{c|}{Modality} & \multicolumn{3}{c|}{Organ} & \multicolumn{3}{c|}{Modality \& Organ} & \multicolumn{3}{c|}{Organ} \\ 
CLIP  && & 44.64   & 45.90  & 50.53 & 94.41 & 92.66 & 90.58& 83.70& 80.38 & 74.31 & 89.11     & 87.66    & 83.96  & 97.08 & 97.05 & 96.18 \\
QwenCLIP &\checkmark& & 45.63 & 46.94 & 51.90 & 96.00 & 94.47 & 92.48 & 85.71 & 83.46 & 78.96 & 91.04 & 90.44 & 88.18 & 98.36 & 97.55 & 96.92 \\
QwenCLIP &\checkmark& \checkmark&\textbf{45.96} & \textbf{47.23} & \textbf{52.47} & \textbf{96.24} & \textbf{94.78} & \textbf{92.61} & \textbf{85.99} & \textbf{83.72} & \textbf{79.22} & \textbf{91.35} & \textbf{90.69} & \textbf{88.42} & \textbf{98.49} & \textbf{97.67} & \textbf{97.34} \\
\hline

\end{tabular}}
\label{tab2}
\end{table*}

\subsection{Results}

In Table~\ref{tab1}, we compare the results obtained with the proposed QwenCLIP framework with baseline models using different text encoders.
Across all benchmarks and metrics, QwenCLIP consistently achieves state-of-the-art performance. On the ROCOv2 dataset, it attains the highest CUI@K, demonstrating superior alignment between image and medical concept (CUI) representations. On our custom retrieval dataset, QwenCLIP also achieves the best P@K scores across all tasks, indicating strong capability in learning discriminative visual features for both anatomical and procedural categories.
Notably, QwenCLIP surpasses LLM2CLIP, which employs a similar 8B-scale language model, highlighting the effectiveness of the Qwen3-embedding backbone and our hybrid prompt tuning strategy in medical contexts. The same performance trend holds on the IRMA dataset, where QwenCLIP achieves the highest P@K scores for organ-based retrieval. These consistent gains across datasets confirm the advantage of leveraging LLM-based embeddings with adaptive prompt tuning for robust medical image–text alignment.

\subsection{Ablation studies}

To validate the contribution of our two core components, we conducted an ablation study (results are summarized in Table~\ref{tab2}). 
Simply replacing CLIP's original text encoder with the Qwen3-Embedding LLM (denoted as "LLM") provides a substantial performance boost over the baseline, confirming the benefit of richer semantic representations. 
The incorporation of our learnable prompt strategy ("LP") further enhances results across all metrics. The full QwenCLIP model, integrating both the LLM encoder and learnable prompts, achieves the best performance, confirming that each component is crucial for optimal cross-modal alignment.


\section{Conclusion}

In this paper, we introduced QwenCLIP, a novel vision-language framework that overcomes the context-length limitations of prior CLIP-based models in medicine. By replacing the standard text encoder with the powerful Qwen3-Embedding LLM, our approach can process long, information-rich clinical reports without truncation. Furthermore, we designed a hybrid prompt tuning strategy to effectively adapt the LLM's semantic space for cross-modal alignment.

Extensive experiments on zero-shot image retrieval benchmarks demonstrate that QwenCLIP achieves state-of-the-art performance, consistently outperforming CLIP, domain-adapted BERT models (ClinicalCLIP, PMC-CLIP), and the recent LLM2CLIP. Ablation studies confirm that both the LLM encoder and our learnable prompts are crucial to this success. While motivated by long-text challenges, QwenCLIP's superior results on datasets with shorter captions (e.g., ROCOv2) highlight its broader advantage: it provides richer semantic representations for medical text of any length, while its architecture is inherently ready to scale to long-form reports—a promising direction for future works and for the community. This study validates the significant potential of dedicated LLM embeddings for advancing medical vision-language understanding.

\section{Compliance with ethical standards}
\label{sec:ethics}
This research study was conducted retrospectively using human subject data made available in open access by medical literature database PubMed Central and Stanford ML Group. Ethical approval was not required as confirmed by the license attached with the open access data.

\section{Acknowledgments}
\label{sec:acknowledgments}

This work was performed using HPC
resources from GENCI-IDRIS (2022-AD011012825R1)
made by GENCI. Xiaoyang Wei is funded by China Scholarship Council.

\bibliographystyle{IEEEbib}
\bibliography{strings,refs}

\end{document}